\documentclass[preprint,12pt]{elsarticle}

\usepackage{amssymb}
 \usepackage{hyperref} 
\usepackage{amsmath}
\usepackage{threeparttable}
\usepackage{longtable}
\usepackage[inkscapelatex=false]{svg}
\usepackage{subfigure}
\usepackage{framed}
\usepackage{footnote}
\usepackage{graphicx}
\usepackage{graphics} 
\usepackage{epsfig}%这一句一定要加上 
\usepackage{xcolor}
\newcommand\gqh[1]{{\color{black}#1}}
\newcommand\zyh[1]{{\color{black}#1}}
\usepackage{xspace}
\usepackage{lipsum}
\usepackage{ulem}

\newcommand{\name}[0]{LLM$^3$-DTI\xspace}
\begin{document}
% % ========== Title Page Start ==========
%   \thispagestyle{empty} 
%   \begin{center}
%   {\Large\textbf{Title Page}}
%   \end{center}

%   \vspace{1cm}  

%   \noindent\textbf{Title:} \\
%   \name: A Large Language Model and Multi-modal data co-powered framework for Drug-Target
%   Interaction prediction

%   \vspace{0.8cm} 

%   \noindent\textbf{Authors:} \\
%   Yuhao Zhang$^{1,4,\dagger}$, Qinghong Guo$^{1,\dagger}$, Qixian Chen$^{2,3,4,*}$, Liuwei
%   Zhang$^{3}$, Hongyan Cui$^{3}$, Xiyi Chen$^{5,*}$

%   \vspace{0.8cm} 

%   \noindent\textbf{Affiliations:}
%   \begin{enumerate}
%   \setlength\itemsep{0.2em}  
%   \item Polytechnic Institute, Zhejiang University, Hangzhou 310015, Zhejiang, China
%   \item School of Pharmaceutical Sciences, Zhejiang University, Hangzhou 310058, Zhejiang,
%   China
%   \item Innovation Center of Yangtze River Delta, Zhejiang University, Jiaxing 314100,
%   Zhejiang, China
%   \item Agricultural Genomics Institute at Shenzhen, Chinese Academy of Agricultural Sciences,
%    Shenzhen 518120, Guangzhou, China
%   \item School of Public Health, Dalian Medical University, Dalian 116044, Liaoning, China
%   \end{enumerate}

%   \vspace{0.5cm}  

%   \noindent$\dagger$ These authors contributed equally to this work.

%   \vspace{0.8cm}  

%   \noindent\textbf{Corresponding authors:}\\

%   \noindent Qixian Chen\\
%   E-mail address: plasmid@zju.edu.cn\\
%   \noindent Xiyi Chen\\
%   E-mail address: xychen@dmu.edu.cn

%   \clearpage 
%   % ========== Title Page End ==========
  
\begin{frontmatter}



\title{\name:  A Large Language Model and Multi-modal data co-powered framework for Drug-Target Interaction prediction}

% -----------作者信息-------------
\author[inst1,inst4]{Yuhao Zhang\fnref{equal}}
\author[inst1]{Qinghong Guo\fnref{equal}}
\author[inst2,inst3,inst4]{Qixian Chen\corref{corresponding}}
\author[inst3]{Liuwei Zhang}
\author[inst3]{Hongyan Cui}
\author[inst5]{Xiyi Chen\corref{corresponding}}

\fntext[equal]{The two authors contribute equally to this work.}
\cortext[corresponding]{Corresponding author.}

\affiliation[inst1]{organization={Polytechnic Institute},%Department and Organization
            addressline={Zhejiang University}, 
            city={Hangzhou},
            postcode={310015}, 
            state={Zhejiang},
            country={China}}

\affiliation[inst2]{organization={School of Pharmaceutical Sciences},%Department and Organization
            addressline={Zhejiang University}, 
            city={Hangzhou},
            postcode={310058}, 
            state={Zhejiang},
            country={China}}

\affiliation[inst3]{organization={Innovation Center of Yangtze River Delta},%Department and Organization
            addressline={Zhejiang University}, 
            city={Jiaxing},
            postcode={314100}, 
            state={Zhejiang},
            country={China}}

\affiliation[inst4]{organization={Agricultural Genomics Institute at Shenzhen},%Department and Organization
            addressline={Chinese Academy of Agricultural Sciences}, 
            city={Shenzhen},
            postcode={518120}, 
            state={Guangzhou},
            country={China}}

\affiliation[inst5]{organization={School of Public Health},%Department and Organization
            addressline={Dalian Medical University}, 
            city={Dalian},
            postcode={116044}, 
            state={Liaoning},
            country={China}}

\begin{abstract}
Drug-target interaction (DTI) prediction is of great significance for drug discovery and drug repurposing. With the accumulation of a large volume of valuable data, data-driven methods have been increasingly harnessed to predict DTIs, reducing costs across various dimensions. 
Therefore, this paper proposes a \textbf{L}arge \textbf{L}anguage \textbf{M}odel and \textbf{M}ulti-\textbf{M}odel data co-powered \textbf{D}rug \textbf{T}arget \textbf{I}nteraction prediction framework, named \name. \name constructs multi-modal data embedding to enhance DTI prediction performance. In this framework, the text semantic embeddings of drugs and targets are encoded by a domain-specific LLM. To effectively align and fuse multi-modal embedding. We propose the dual cross-attention mechanism and the TSFusion module. Finally, these multi-modal data are utilized for the DTI task through an output network. The experimental results indicate that \name can proficiently identify validated DTIs, surpassing the performance of the models employed for comparison across diverse scenarios. Consequently, \name is adept at fulfilling the task of DTI prediction with excellence. The data and code are available at \href{https://github.com/chaser-gua/LLM3DTI}{{https://github.com/chaser-gua/LLM3DTI.}}
\end{abstract}


\end{frontmatter}


\section{Introduction}
\label{sec:sample1}
The development of targeted drugs for various diseases is regarded as an essential and effective approach in modern medicine~\cite{jiang2024reviewjpa}. Nonetheless, estimates suggest that it takes at least 20 years and \$2 billion to bring a Food and Drug Administration (FDA) approved drug from initial biological screening and development to postmarket testing, significantly hindering the implementation of precision therapies~\cite{schlander2021}. Consequently, drug repurposing, which explores the reuse potential of existing drugs, has emerged as a strategy to accelerate the development of targeted therapies~\cite{HU2024101084jpa}. The core of drug repurposing lies in predicting and identifying potential drug-target interactions (DTIs)~\cite{xu2021}. Currently, computational methods are widely employed for DTI prediction, effectively addressing the high costs and time-consuming nature of biochemical experiments~\cite{ding2020}. Computational DTI models can be broadly categorized into two types: docking-based models and data-driven models.

Docking-based approaches play a crucial role in drug development and biochemistry~\cite{bhargava2021}. These approaches rely on the fundamental assumption that ligands with similar chemical properties typically exhibit similar biological activities and can bind to similar target proteins. Interactions between new ligands and proteins are predicted by leveraging structural information from known active ligands through structural similarity comparisons. However, if the number of known ligands binding to a specific target protein is too limited, the predictive reliability of such methods may be compromised~\cite{WANG2024101134jpa}.

Data-driven methods, central to machine learning and deep learning, extract and utilize latent information from drug and target data~\cite{FENG2023115107jpa}. Early approaches focus on manual construction of drug and target features integrated with machine learning models, such as support vector machines~\cite{basak2007support} (SVM) and random forests~\cite{liaw2002classification} (RF). While manually engineered features incorporate comprehensive prior knowledge, they fail to capture intricate nonlinear relationships and high-dimensional data patterns inherent in biological systems~\cite{ZHANG2024101159jpa}, limiting their practical application.
With the accumulation of large-scale genomic and proteomic data and the advancement of deep neural network technologies, deep learning models have emerged as new methods for DTI prediction~\cite{CHEN2024101161jpa}. 
Initially, researchers~\cite{monteiro2020, fernandez2022, 10.1371/journal.pone.0307649} independently encode drug and target features, containing the Simplified Molecular Input Line Entry System (SMILES) format for drug chemical structures and molecular descriptors or sequences for targets. 
Although these models are structurally and conceptually simple, they rely exclusively on static molecular features, neglecting network topology within interaction networks. Their separate encoding of drugs and targets overlooks interactive characteristics.

In contrast, graph-based models enhance prediction accuracy by integrating network connectivity between drugs and targets~\cite{zeng2020}. Some efforts focus on mining latent semantic information from graph structures, framing the DTI prediction problem as a link prediction task~\cite{ye2021}. For instance, Muhammad et al.~\cite{Mahmud2021} develop a graph-based model that integrates drugs, targets, and related entities into a knowledge graph, computing drug-target matching scores using graph embedding techniques. Ye et al.~\cite{ye2021} introduce recommendation system techniques to derive low-dimensional representations of drugs and targets from knowledge graphs. These methods effectively leverage structural information from the knowledge graph, but they heavily depend on the completeness of graphs, with missing links potentially biasing predictions. Others improve prediction accuracy by learning the topological features of drug-target networks. For example, Li et al.~\cite{Li2022} propose the HGAN-DTI model, which utilizes a heterogeneous graph attention network to capture information transfer between non directly connected nodes, thereby establishing topological features for drugs and targets. Yuan et al.~\cite{Yuan2023} introduce the EDC-DTI model, employing an enhanced graph attention mechanism to integrate various entity features of drugs and targets, capturing multi-scale topological features. Graph-based methods have advanced through network topology incorporation. However, they fail to integrate textual modality information, such as the description of \zyh{drugs} and \zyh{targets} in biological literature or databases, which provides valuable contextual knowledge for DTI prediction. 
Consequently, there are several key issues that require resolution. First, textual mining for drugs and targets remains superficial; deeper extraction of semantic information from textual modalities is essential. Second, structural topology network features and textual features constitute multi-modal data, necessitating effective alignment and fusion strategies to enhance DTI prediction performance.

To address these challenges, we propose a \textbf{L}arge \textbf{L}anguage \textbf{M}odel and \textbf{M}ulti-\textbf{M}odel data co-powered \textbf{D}rug \textbf{T}arget \textbf{I}nteraction prediction framework (\name).
For the first challenge, we collect textual descriptions of drugs and targets sourced from public databases and employ pharmaceutical-domain fine-tuned large language models to encode them for comprehensive semantic information. With the rapid advancement of large language models (LLMs), these cutting-edge technologies are increasingly being applied to deep learning. LLMs such as LLaMA~\cite{Touvron2023}, which are trained on extensive datasets and fine-tuned for domain-specific applications, provide a foundation for comprehensively utilizing textual modality information in DTI prediction. To the best of our knowledge, we are the first to leverage the power of LLMs to aid DTI prediction.
For the second challenge, we design a dual cross-attention mechanism and a textual and structural topology modality fusion module (TSFusion) to effectively align and fuse the multi-modal data. Specifically, we employ the dual cross-attention mechanism to replace self-attention computation, facilitating the complementation of textual and structural modalities. The TSFusion employs an adaptive gating mechanism to dynamically adjust inter-modal weight distributions, refining the integration of textual and structural topological embeddings through precise importance balancing. More details of \name are provided in section~\ref{sec:sample:2}.
Extensive experimental results and visualization analyses demonstrate that our method surpasses existing methods.
In summary, the main contributions of our work can be summarized as follows:
\begin{itemize}
\item[1)] 
We propose \name, \zyh{an} LLM and multi-modal data co-powered DTI prediction framework, employing a domain-specific LLM to encode textual drug and target descriptions. 
% To our knowledge, it is the first application of domain-specific LLM in DTI prediction.
\item[2)]
We design a dual cross-attention mechanism and a TSFusion module to align and fuse multi-modal data. These components enhance multi-modal complementarity.
\item[3)] We conduct a variety of experimental tasks for \name. Extensive experimental results show that our method is superior to other models in prediction accuracy and robustness.
\end{itemize}

\section{Material and methods}\label{sec:sample:2}
\begin{figure}[bp!]
\centering
{\includegraphics[width=\linewidth]{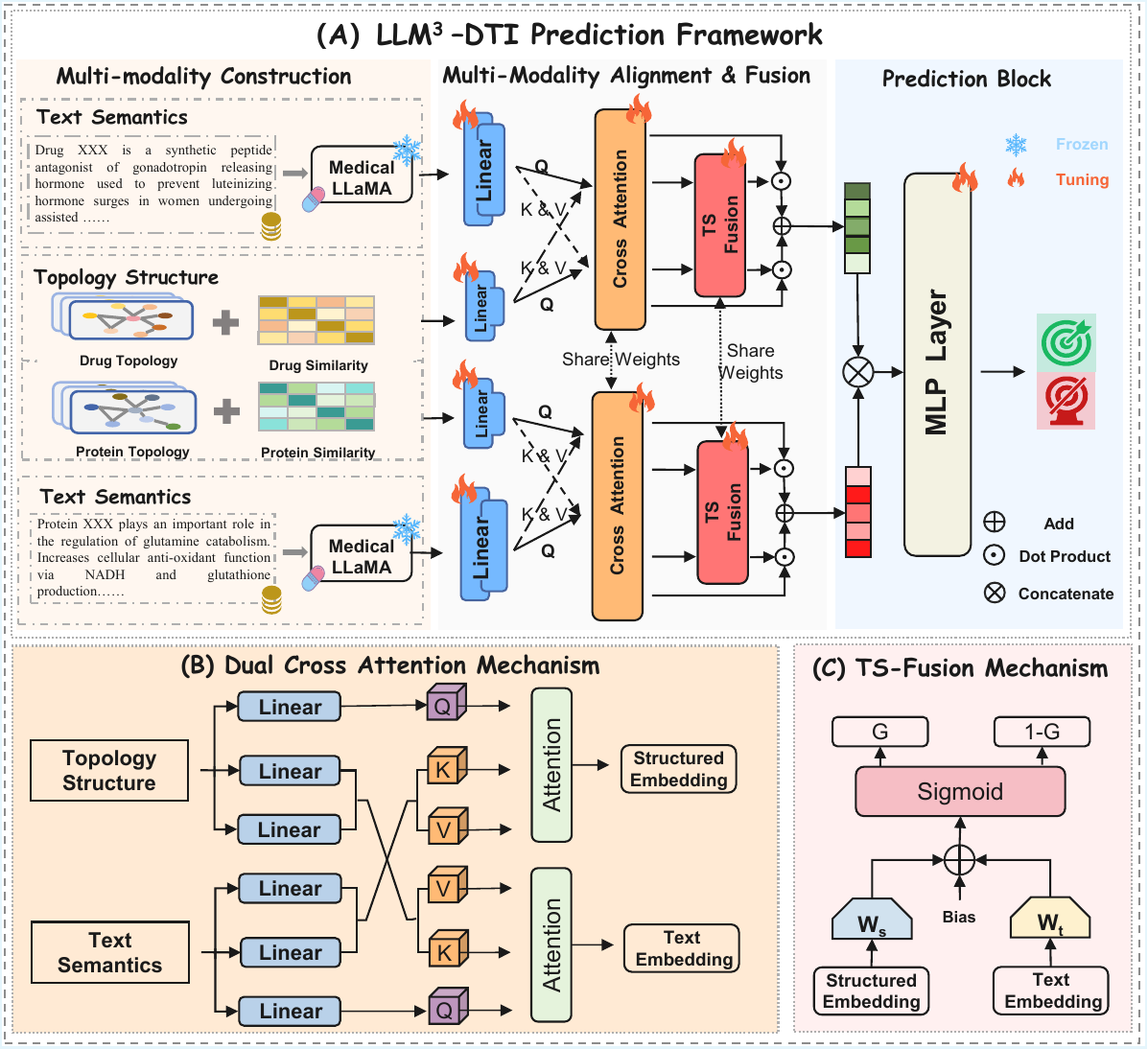}}
\caption{The overall framework we proposed. }
\label{fig1}
\end{figure}

Figure~\ref{fig1} presents the overall architecture of \name. (A) depicts the pipeline of \name, while (B) and (C) illustrate the detailed mechanisms of dual cross-attention and the TSFusion module, respectively. Specifically, \name employs multi-modal data to enhance DTI prediction performance. This approach comprises four key components: multi-modal embedding construction for drug and target topology and text descriptions; the dual cross-attention module for alignment across modality data; the TSFusion module for cross-modality data fusion; and the DTI prediction block. The subsequent section details each module.

% \begin{table}[htbp]
% \centering
% \caption{Notations used in \name.}
% \resizebox{\columnwidth}{!}{
% \begin{tabular}{ll}
% \hline
% \textbf{Symbol} & \textbf{Definition} \\ \hline

% \label{notation}
% \end{table}

\subsection{Multi-modal Embedding Construction}
As previously stated, \name primarily utilizes two modal data: the structural topology data aggregating entities, and textual data describing mechanisms of action from the DrugBank and UniProt databases. This section details the construction of embeddings for both drug and target data.

\subsubsection{Structural topology embedding}
% Inspired by Luo et. al.~\cite {Luo2017}, for structural features, we mainly consider the similarity of homogeneous information and the features from heterogeneous information. Drug and target similarity-based features are extracted from drug-drug and protein-protein association networks by computing the Jaccard similarity~\cite{jaccard1912distribution} among entities in these networks.
Inspired by Luo et al~\cite{Luo2017}, we consider homogeneous similarity information and heterogeneous graph network information for structural topology data. Similarity-based features for drugs and targets are extracted from drug-drug and protein-protein association networks through Jaccard similarity computation among network entities. Graph-based features from heterogeneous interaction networks are computed using graph topology algorithms and eigenvalue decomposition. Specifically, the Random Walk with Restart (RWR) algorithm calculates graph topology, while Diffusion Component Analysis (DCA) reduces dimensionality. Three structural network types are utilized: drug-disease association networks, drug-side effect association networks, and protein-disease association networks. Within each network, RWR executes independently, generating diffusion state vectors for every node. After obtaining diffusion state vectors, DCA reduces the high-dimensional data into compact low-dimensional representations. This step employs eigenvalue decomposition to extract global network information. Collectively, these eigenvectors and eigenvalues capture the most significant structural topological information of the graph networks. At this stage, the structural topological embeddings of the drug $Z^d_s \in \mathbb{R}^{N_d \times d_1}$ and target $Z^p_s \in \mathbb{R}^{N_p \times d_2}$ have been derived, where $N_d$ and $N_p$ are the number of drugs and targets; $d_1$ and $d_2$ denote the number of dimensions, respectively.

\subsubsection{Text semantic embedding}

In contrast to employing language models (e.g., Bert~\cite{devlin2019bert}) for encoding drug SMILES strings and target amino acid sequences, we introduce textual descriptions of drugs and targets. Specifically, drug summary and mechanism of action fields are extracted from DrugBank, while target function text is obtained from UniProt. This raw text undergoes preprocessing to remove extraneous elements, such as HTML tags, special characters, and comments.
% \gqh{In the preprocessing phase, we harmonized identifiers across different databases. For drug entities, DrugBank IDs served as primary identifiers, supplemented by mappings through InChI Key or SMILES structures where necessary. Protein entities were standardized using UniProt accession numbers. We further addressed the issue of synonyms and aliases by consolidating terms through biomedical lexicons such as MeSH or UMLS, ensuring uniformity and minimizing redundancy.Entries lacking essential information or identifiers were removed, while numerical fields with occasional missing values were imputed using median or mean values. Duplicate records, especially in drug-target pairs or network connections, were rigorously identified and consolidated, thereby preventing any disproportionate influence during model training.Raw texts underwent removal of irrelevant content such as HTML tags, special characters, and annotations. Additionally, text segments were logically divided into meaningful prompts designed for LLM-based embedding generation. We constructed both homogeneous and heterogeneous interaction networks, carefully filtering edges based on evidence strength and reliability scores from respective databases. Redundant or conflicting edges were systematically identified and resolved either by merging reliable connections or discarding those with insufficient support.}

Leveraging technological advances from large language models, we employ Medical-LLaMa, a biomedical LLM fine-tuned on domain-specific data, to generate contextual text embeddings. The primary motivation behind our design is to effectively extract meaningful semantic information from complex biomedical textual descriptions of drugs and proteins.
For both drugs and proteins, embeddings are extracted from the last hidden layers of Medical-LLaMA, representing high-dimensional semantic information. At this stage, the text semantic embeddings of the drug $Z^d_t \in \mathbb{R}^{N_d \times d_3}$ and target $Z^p_t \in \mathbb{R}^{N_p \times d_3}$ have been derived, where $d_3$ denotes the number of dimensions.
% These embeddings, denoted $ {F}_{\text{LLM,d}} $ and ${F}_{\text{LLM,p}} $, are formally defined as:

% \begin{align}
% {F}_{\text{LLM},d} &\in {R}^H \\
% {F}_{\text{LLM},p} &\in {R}^H
% \end{align}
% Let ${F}_{\text{LLM},d} $ refer to the embedding representation for drug textual features and $ {F}_{\text{LLM},p} $ refer to the embedding representation for protein textual features. The variable $H$ denotes the dimension of the final hidden states in the LLM. These unstructured embeddings enrich the feature representations by capturing biochemical contexts and functional properties for drugs and proteins.

\subsection{Dual Cross-attention Alignment}
To align embeddings across modalities, we design a dual cross-attention alignment module. The cross-attention mechanism effectively integrates complementary information from different modalities, enhancing multi-modal embedding alignment~\cite{yan2024urbanclip}. Before cross-attention computation, drug and target multi-modal embeddings are projected to identical dimensions through two linear layers. When updating structural topological features via dual cross-attention, these features serve as queries while corresponding textual features act as keys and values. Conversely, during text feature updates, textual features function as queries, and structural topological features serve as keys and values. The drug and target multi-modal embedding alignment process operates as follows:
% To update the drug features, the embedded target features \({Z}_{p}\) are used as the query, while the embedded drug features \({Z}_{d}\) serve as the key and value. The updated drug features \({F}_{\text{aligned},d}\) are computed as:
\begin{align}
{Z}^d_{s-cra} = \text{Softmax}\left(\frac{{Q}^d_s {K^d_t}^\top}{\sqrt{d_k}}\right){V}^d_t , 
{Z}^d_{t-cra} = \text{Softmax}\left(\frac{{Q}^d_t {K^d_s}^\top}{\sqrt{d_k}}\right){V}^d_s ,
\\
{Z}^p_{s-cra} = \text{Softmax}\left(\frac{{Q}^p_s {K^p_t}^\top}{\sqrt{d_k}}\right){V}^p_t , 
{Z}^p_{t-cra} = \text{Softmax}\left(\frac{{Q}^p_t {K^p_s}^\top}{\sqrt{d_k}}\right){V}^p_s ,
\end{align}

\begin{align}
\begin{aligned}
{Q}^d_s, {Q}^d_t, {Q}^p_s, {Q}^p_t = {Z}^d_{s} W_{q}, {Z}^d_{t} W_{q}, {Z}^p_{s} W_{q}, {Z}^p_{t} W_{q},  \\
{K}^d_s, {K}^d_t, {K}^p_s, {K}^p_t = {Z}^d_{t} W_{k}, {Z}^d_{s} W_{k}, {Z}^p_{t} W_{k}, {Z}^p_{s} W_{k},  \\
{V}^d_s, {V}^d_t, {V}^p_s, {V}^p_t = {Z}^d_{t} W_{v}, {Z}^d_{s} W_{v}, {Z}^p_{t} W_{v}, {Z}^p_{s} W_{v},
\end{aligned}
\end{align}
where \(W_{q}\), \(W_{k}\), and \(W_{v}\) are learnable weight matrices, and \(d_k\) represents the dimensionality of the key, used for scaling. To enhance drug-target information interaction during forward propagation, the drug and target cross-attention modules share parameter weights.

% To update the target features, the embedded drug features \({Z}_{d}\) are used as the query, while the embedded target features \({Z}_{p}\) serve as the key and value. The updated target features \({F}_{\text{aligned}, p}\) are computed as:  

% \begin{equation}
% {F}_{\text{aligned},p} = \text{softmax}\left(\frac{{Q}_d {K}_p^\top}{\sqrt{d_k}}\right){V}_p,
% \end{equation}

% \begin{align}\left\{\begin{aligned}
% {Q}_d = {Z}_{d} W_{q,p},\\
% {K}_p = {Z}_{p} W_{k,p},\\
% {V}_p = {Z}_{p} W_{v,p},
% \end{aligned}\right.\end{align}
% where \(W_{q,p}\), \(W_{k,p}\), and \(W_{v,p}\) are learnable weight matrices.

% Following the updates through the cross attention modules, the updated drug features \({F}_{\text{aligned}, d}\) and target features \({F}_{\text{aligned}, p}\) are concatenated to form the joint drug-target feature representation \({F}_{\text{joint}}\), as follows:  

% \begin{equation}
% {F}_{\text{joint}} = [{F}_{\text{aligned}, d}, {F}_{\text{aligned}, p}].
% \end{equation} 

% The final feature representation \({F}_{\text{joint}}\) is subsequently utilized in the prediction network to infer drug-target binding. The feature alignment module leverages the complementary information from multi-modal features, producing a more expressive joint representation and establishing a robust foundation for subsequent prediction tasks.

\subsection{Text and Structure embedding Fusion}
Structural topology and textual semantic features of drug or target entities capture complementary characteristics, necessitating fusion for comprehensive representations. However, simple addition or static weighting may induce modal conflicts~\cite{wei2025path}. To enhance multi-modal fusion for subsequent DTI prediction, we introduce a text-semantic and structural-topology fusion module, named TSFusion. This module dynamically adjusts modal contributions through location-selective weighting, improving fusion accuracy. Specifically, linear transformations assign weights to both modality features, with the fused output being their weighted sum. The TSFusion formula is:
\begin{align}
\begin{aligned}
    G^d = \text{Sigmoid}\left(W_sZ^d_{s-cra} + W_tZ^d_{t-cra} + b\right), \\
    G^p = \text{Sigmoid}\left(W_sZ^p_{s-cra} + W_tZ^p_{t-cra} + b\right),
\end{aligned}
\end{align}
\begin{align}
\begin{aligned}
   Z^d_{fusion} = G^d \cdot Z^d_{s-cra} + \left(1-G^d)\right. \cdot Z^d_{t-cra},\\
   Z^p_{fusion} = G^p \cdot Z^p_{s-cra} + \left(1-G^p)\right. \cdot Z^p_{t-cra},
\end{aligned}
\end{align}
where $W_s$ and $W_t$ are learnable weight matrices, $b$ is the bias term, \zyh{and} \text{Sigmoid} is the activation function \zyh{that maps} the $G$ into the range of 0 and 1. Similarly, drug multi-modal feature fusion and target multi-modal feature fusion utilize identical TSFusion weights.

\subsection{Prediction block and loss function}

% In the final prediction stage, the drug and target representations derived from preceding stages are concatenated for final DTI prediction. The prediction block maps the final representation to interaction probabilities through fully connected layers followed by activation functions. The drug-target binding probability $\hat{y}$ is calculated as:
In the final prediction stage, drug and target representations from preceding stages are concatenated for DTI prediction. The prediction block transforms this concatenated representation into interaction probabilities using fully connected layers and activation functions. The detailed computed formulas are:
\begin{equation}
Z = Z^d_{fusion} \vert\vert Z^p_{fusion} ,
\end{equation}

\begin{equation}
\hat{y} = \sigma({w}^\top {Z} + b),
\end{equation}
where $\vert\vert$ denotes the concatenate operation, $w$ represents the trainable weights of the prediction block, $Z$ is the joint representation of drug and protein features, $b$ denotes the bias term, and $\sigma$ is the activation function.

% \begin{equation}
% \sigma(x) = \frac{1}{1 + e^{-x}}.
% \end{equation}
% Here, $x$ is the input to the Sigmoid function. The sigmoid function maps the input values to a probability range of $(0, 1)$.

% To optimize the binary classification performance of the model, the Binary Cross Entropy (BCE) loss function is used. The loss function $\mathcal{L}_{\text{BCE}}$ is defined as:
Consistent with established methodologies, binary cross-entropy loss (BCE) optimizes all framework parameters. The loss is computed as follows:

\begin{equation}
\mathcal{L}_{\text{BCE}} = -\frac{1}{N} \sum_{i=1}^{N} \left[ y_i \log(\hat{y}_i) + (1 - y_i) \log(1 - \hat{y}_i) \right],
\end{equation}
where $N$ is the total number of samples, $y_i$ denotes the ground truth label for the $i$-th sample, and $\hat{y}_i$ is the predicted probability for the $i$-th sample. This loss function ensures accurate DTI predictions by penalizing incorrect predictions for both positive and negative samples.

\section{Experiment Results}

This section details the experimental setup and results. The findings indicate that the proposed \name framework achieves state-of-the-art performance, demonstrating that properly aligned and fused multi-modal data significantly enhances DTI prediction accuracy.

\subsection{Datasets}
% The dataset, referred to as LuoData, is utilized by DTINET. It encompasses 708 drugs, 1,493 targets and additional heterogeneous information such as disease and side effects. These data are predominantly sourced from authoritative databases such as DrugBank, HPRD, CTD, SIDER and UniProt. The dataset includes comprehensive and credible descriptions of molecular characteristics, interactions, toxicity and safety, thereby providing robust support for model training and evaluation. Due to its wide acceptance and significant credibility in DTI tasks, LuoData is selected as the training and evaluation dataset for the \name model in this study.

% The dataset, denoted as LuoData, is employed in DTINET. It comprises 708 drugs, 1,493 targets, and heterogeneous data (e.g., disease associations, side effects) primarily sourced from authoritative databases including DrugBank, HPRD, CTD, SIDER, and UniProt. LuoData provides detailed descriptors of molecular characteristics, interaction profiles, toxicity, and safety parameters, serving as a reliable foundation for model training and evaluation. Owing to its broad recognition and high credibility in DTI research, this study adopts LuoData as the training and evaluation dataset for the \name framework.

The dataset we used contains 708 drugs, 1,493 targets, and heterogeneous data, including disease associations and side effects—primarily sourced from authoritative databases such as DrugBank, HPRD, CTD, SIDER, and UniProt. This data is usually used by previous works, offering comprehensive descriptors of molecular characteristics, interaction profiles, toxicity, and safety parameters, providing a reliable foundation for model training and evaluation. Given the broad recognition and high credibility of the data in DTI research, this dataset serves as the training and evaluation resource for the \name framework. To rigorously assess the performance of the \name model, we split the dataset into three distinct subsets for training, validation, and independent testing. The independent test set was exclusively held out and not used in any model training or hyperparameter tuning phase.

\subsection{Baselines}
To evaluate the performance of the proposed framework, \name is compared against strong models for DTI prediction.

% SVM~\cite{basak2007support} is a supervised learning based model for DTI prediction, designed to classify data from different categories by constructing an optimal hyperplane. 
SVM~\cite{basak2007support} is a supervised learning model for DTI prediction that classifies data into distinct categories by constructing an optimal hyperplane.

% RF~\cite{liaw2002classification} is an ensemble learning-based model for DTI prediction, achieved by integrating predictions across multiple decision trees. This method has been widely used in various domains.
RF~\cite{liaw2002classification} is an ensemble learning model for DTI prediction that integrates predictions from multiple decision trees. 

% DTINet~\cite{Luo2017} is a network integration framework aimed at exploiting heterogeneous biological data sources DTI prediction. DTINet learns low-dimensional, information-rich feature vectors for drugs and proteins.
DTINet~\cite{Luo2017} is a network integration framework that leverages heterogeneous biological data sources for DTI prediction. This approach provides information-rich feature representations for drugs and proteins.

% GCN-DTI~\cite{Shao2022} is a DTI prediction model that utilizes a graph convolutional network. The model applies graph convolution operations to extract features from molecular graphs or biological networks that describe drugs and targets, with adjacency matrices used to represent node connections. 
GCN-DTI~\cite{Shao2022} is a DTI prediction model employing graph convolutional networks. It extracts features from molecular graphs or biological networks representing drugs and targets through graph convolution operations.

% GAT-DTI~\cite{Wang2021} is a model for DTI prediction that leverages the GAT. Using the multi-head attention mechanism of GAT, the model extracts features from drug and target data by representing their topological relationships using adjacency matrices. 
GAT-DTI~\cite{Wang2021} is a DTI prediction model utilizing graph attention networks (GAT). It extracts features from drug and target data through the multi-head attention mechanism of GAT.

% DTI-CNN~\cite{peng2020} is a deep convolutional neural network (CNN) based model for DTI prediction. This method uses three main components: a heterogeneous network feature extractor, a denoising autoencoder-based feature selector and a CNN interaction predictor to predict.
DTI-CNN~\cite{peng2020} enhances performance through three core components: a heterogeneous network feature extractor, a denoising autoencoder feature selector, and a convolutional neural network-based interaction predictor.

% HHDTI~\cite{Ruan2021} captures high-order relations within heterogeneous biological networks to improve predictive capability. This method uses a hypergraph structure to model the intricate topological relationships, generating embedding representations for DTI prediction.
% HHDTI~\cite{Ruan2021} employs hypergraph structures to model high-order relationships within heterogeneous biological networks, generating embedding representations for DTI prediction.

% IMCHGAN~\cite{Li2022} is an end-to-end neural network learning framework that employs a two-stage attention mechanism to learn the respective features of drugs and targets from heterogeneous networks. The model leverages a local attention mechanism to focus on the key features of drugs or targets and a global attention mechanism to capture the potential interaction relationships between drugs and targets. 
IMCHGAN~\cite{Li2022} employs a two-stage attention mechanism to extract drug and target features from heterogeneous networks. This approach utilizes a local attention mechanism to identify key drug or target features and a global attention mechanism to capture potential drug-target interactions.

% DTI-LM~\cite{Ahmed2024} leverages language models and GAT to generate rich encoding representations from protein amino acid sequences and drug SMILES sequences. This method uses simple text semantic information for DTI prediction.
DTI-LM~\cite{Ahmed2024} employs language models and GAT to generate rich encoded representations from protein amino acid sequences and drug SMILES strings.
% This approach utilizes textual semantic information for drug-target interaction (DTI) prediction.

% CCL-ASPS~\cite{Tian2024} is a deep learning model designed specifically for DTI prediction, integrating collaborative contrastive learning and adaptive self-paced sampling techniques. These two techniques can improve the ability of the model to capture the subtle interaction differences and help the model achieve better performance.
CCL-ASPS~\cite{Tian2024} integrates collaborative contrastive learning and adaptive self-paced sampling techniques to enhance the capture of subtle interaction differences and improve model performance.

\subsection{Evaluation metrics}
% In this study, multiple evaluation metrics are selected to assess the performance of deep learning based models for DTI prediction. These metrics include Accuracy (ACC), the Area Under the Receiver Operating Characteristic Curve (AUROC), the Area Under the Precision-Recall Curve (AUPR), the Matthews Correlation Coefficient (MCC) and the F1 score. The detailed description and definition of each metric are as follows:

This study employs five evaluation metrics to comprehensively assess the performance of \name and compared methods in DTI prediction: Accuracy (ACC), Area Under the Receiver Operating Characteristic Curve (AUROC), Area Under the Precision-Recall Curve (AUPR), Matthews Correlation Coefficient (MCC), and F1 Score. The detailed formula of ACC is:

\begin{equation}
\text{ACC} = \frac{\text{TP} + \text{TN}}{\text{TP} + \text{TN} + \text{FP} + \text{FN}},
\end{equation}
where TP denotes the number of instances correctly predicted as positive, TN represents the number of instances correctly predicted as negative, FP refers to the number of incorrect predictions as positive, and FN represents the number of incorrect predictions as negative. ACC is one of the most fundamental metrics for evaluating the performance of a classification model, measuring the proportion of correctly classified predictions. However, it may lead to biased evaluations in the presence of imbalanced datasets when used as a standalone metric. Therefore, we consider the introduction of AUPR and F1 score for further model evaluation. The computation is:

\begin{equation}
\text{Precision} = \frac{\text{TP}}{\text{TP} + \text{FP}}, \quad \text{Recall} = \frac{\text{TP}}{\text{TP} + \text{FN}},
\end{equation}

\begin{equation}
\text{F1} = 2 \cdot \frac{\text{Precision} \cdot \text{Recall}}{\text{Precision} + \text{Recall}}.
\end{equation}
AUPR assesses the performance for the positive class by computing the area under the precision-recall curve. F1 Score is the harmonic mean of precision and recall, balancing the performance of a model between precision (positive predictive value) and recall (sensitivity). They are suitable in the case of imbalanced positive and negative samples, because precision and recall are taken into account in the calculation of these two indices:

\begin{equation}
\text{MCC} = \frac{\text{TP} \cdot \text{TN} - \text{FP} \cdot \text{FN}}{\sqrt{(\text{TP} + \text{FP}) (\text{TP} + \text{FN}) (\text{TN} + \text{FP}) (\text{TN} + \text{FN})}},
\end{equation}

\begin{equation}
\text{TPR} = \frac{\text{TP}}{\text{TP} + \text{FN}}, \quad \text{FPR} = \frac{\text{FP}}{\text{FP} + \text{TN}}.
\end{equation}
MCC is a metric that comprehensively considers the classification accuracy for both positive and negative samples. AUROC evaluates the ability to distinguish between positive and negative samples. The ROC curve is constructed by varying the classification threshold, plotting the relationship between the TPR and the FPR.

\subsection{Experiment settings}  
% \gqh{(TBD)} 
% After a thorough hyperparameter search and optimization process, appropriate hyperparameters are configured to accelerate the convergence of the deep learning model and enhance its performance. In the experiments, the output dimensionality of the vector after dimensional reduction is set to 256, the hidden layer dimensionality in the cross attention module is set to 512, the number of attention heads is specified as 4, the dropout rate is set to 0.2, and the maximum number of epochs is fixed at 500. The AdamW optimizer is selected due to its superior handling of weight decay, which mitigates unnecessary impacts on bias parameters and improves model training effectiveness. The learning rate and weight decay rate of the AdamW optimizer are set to 1e-3 and 1e-6, respectively, to optimize the model parameters efficiently.

Appropriate hyperparameters can accelerate convergence and enhance deep learning model performance. Table~\ref{tab:hs} details some key hyperparameter configurations for \name. For more analysis on parameter sensitivity, please refer to section~\ref{sec:pa}.

% \begin{table}[bp!]
% \centering
% \begin{tabular}{|c|c|c|c|c|c|c|c|}
% \hline
% Hyperparameter & Training epochs &  Hidden layer dimension & Optimizer & Learning rate & Weight decay & Batch size & Layer of prediction block  \\
% \hline
% Setting & 100 & 128 & AdamW & 1e-3 & 1e-6 & 64 & 2 \\
% \hline
% \end{tabular}
% \caption{Hyperparameter settings.}
% \label{tab:hs}
% \end{table}

\begin{table}[tp!]

\centering
\caption{Hyperparameter settings.}
\begin{tabular}{|c|c|} 
\hline
\textbf{Hyperparameter} & \textbf{Setting} \\ 
\hline
Training epochs & 100 \\
\hline
Hidden layer dimension & 128 \\
\hline
Optimizer & AdamW \\
\hline
Learning rate & 1e-3 \\
\hline
Weight decay & 1e-6 \\
\hline
Batch size & 64 \\
\hline
Layer of prediction block & 2 \\
\hline
\end{tabular}

\label{tab:hs}
\end{table}

% Following rigorous hyperparameter optimization, key configurations are selected to optimize model convergence and performance. The experiments employ the following settings: dimensionality-reduced output vectors of size 256, cross-attention module hidden layers of size 512, 4 attention heads, a dropout rate of 0.2, and a maximum training duration of 500 epochs. The AdamW optimizer is adopted for its effective weight decay regularization, which minimizes unintended bias parameter interference while enhancing training efficiency. Learning and weight decay rates are set to 1e-3 and 1e-6, respectively, to ensure stable parameter optimization.

\subsection{Analysis of performance}
% The proposed method is comprehensively compared with the aforementioned machine learning and deep learning approaches across five evaluation metrics. To ensure a fair comparison, the parameter settings described in the original implementations of these methods were strictly followed. All methods were evaluated using five-fold cross-validation, with experiments conducted on five different random seeds. Specifically, for each seed, all positive samples from the dataset were first extracted, and negative samples were randomly generated in equal proportions to create the complete sample set. During each iteration, the sample set was randomly split into five subsets, with four subsets used for training and one subset reserved for testing. The mean values of the evaluation metrics obtained across the five cross validation folds were reported as the experimental results for each seed. The model's final performance was subsequently expressed as the mean and standard deviation of the results across the five random seeds. This approach provides a comprehensive assessment of the model's ability to classify samples and minimizes the impact of random errors on the evaluation process.
% \gqh{TBD}

We compare the \name with the aforementioned machine learning and deep learning methods using five evaluation metrics. To ensure fairness, the hyperparameter configurations for baseline methods adhere strictly to their original implementations. All models undergo repeated training across five different random seeds. For each seed, positive samples are extracted from the dataset, with negative samples generated at a 1:1 ratio to form balanced datasets. The final model performance is reported as the mean and standard deviation across all five seeds. This rigorous evaluation ensures classification robustness while mitigating bias from random variations.

The results are summarized in Table~\ref{tab1}, with the best performance values for each metric highlighted in bold and second-best results underlined. It can be observed that the proposed model consistently outperforms all baseline methods across all five evaluation metrics, confirming superior performance. First, traditional machine learning methods such as SVM and RF cannot fully model nonlinear relationships in drug-target characterization, resulting in the lowest performance. Second, GCN-DTI and GAT-DTI utilize network topology to consider interactions between entities, improving performance over traditional methods. However, these approaches fail to leverage heterogeneous and multi-modal data, yielding suboptimal results. Third, while methods such as IMCHGAN, DTI-LM, and CCL-ASPS further enhance performance by exploiting heterogeneous, multi-modal data and multi-view graph topology, they lack a comprehensive design for multi-modal data fusion. Finally, \name introduces comprehensive textual modality, realizing multi-modal data contributions to DTI prediction through a well-designed alignment and fusion mechanism. Specifically, compared to the second-best baseline, \name achieves improvements of 2.17\% in ACC, 2.32\% in F1-score, 0.4\% in AUPR, 3.26\% in MCC, and 0.4\% in AUROC. Notably, \name demonstrates significant gains over DTI-LM, which also incorporates textual semantics. This underscores the importance of capturing rich semantic representations from textual data and highlights the superior capabilities of LLM for semantic understanding and encoding. These findings provide valuable insights for future development of DTI approaches.

\begin{table}[tp!]
\centering
% \scalebox{0.5}{
\caption{Performance comparison of models for DTI prediction. Results are presented as mean ± standard deviation across five random seeds. The best and second-best values per metric are highlighted in bold and underlined, respectively. An asterisk * indicates statistical significance (p $\textless$ 0.05).}
\resizebox{\columnwidth}{!}{
\begin{tabular}{lccccc}
\hline
 Model & ACC & F1 & AUPR & MCC & AUROC \\
\hline
SVM & ${0.5320\pm0.009}^{*}$ & ${0.6661\pm0.016}^{*}$ & ${0.7148\pm0.032}^{*}$ & ${0.1146\pm0.029}^{*}$ & ${0.7030\pm0.027}^{*}$ \\
RF & ${0.7686\pm0.006}^{*}$ & ${0.7363\pm0.009}^{*}$ & ${0.8617\pm0.006}^{*}$ & ${0.5544\pm0.011}^{*}$ & ${0.8413\pm0.006}^{*}$ \\
DTINET & ${0.5135\pm0.001}^{*}$ & ${0.0528\pm0.002}^{*}$ & ${0.9051\pm0.002}^{*}$ & ${0.1164\pm0.003}^{*}$ & ${0.8730\pm0.002}^{*}$ \\
GCN-DTI & ${0.6145\pm0.002}^{*}$ & ${0.7218\pm0.001}^{*}$ & ${0.6036\pm0.007}^{*}$ & ${0.3593\pm0.004}^{*}$ & ${0.5849\pm0.012}^{*}$ \\
GAT-DTI & ${0.6247\pm0.003}^{*}$ & ${0.7325\pm0.013}^{*}$ & ${0.6620\pm0.059}^{*}$ & ${0.3699\pm0.004}^{*}$ & ${0.6348\pm0.015}^{*}$ \\
DTI-CNN & ${0.8601\pm0.002}^{*}$ & ${0.8605\pm0.003}^{*}$ & ${0.9384\pm0.002}$ & ${0.7373\pm0.005}$ & ${0.9258\pm0.003}$ \\
IMHGAN & ${\underline{0.8680\pm0.016}}$ & ${\underline{0.8614\pm0.017}}$ & ${\underline{0.9410\pm0.005}}$ & ${\underline{0.7392\pm0.031}}$ & ${\underline{0.9350\pm0.006}}$ \\
DTI-LM & ${0.7821\pm0.005}^{*}$ & ${0.7838\pm0.005}^{*}$ & ${0.8790\pm0.004}^{*}$ & ${0.5654\pm0.010}^{*}$ & ${0.8667\pm0.004}^{*}$ \\
CCL-ASPS & ${0.8656\pm0.006}$ & ${0.8623\pm0.006}$ & ${0.9343\pm0.007}^{*}$ & ${0.7343\pm0.012}$ & ${0.9303\pm0.005}$ \\
\textbf{Ours} & \textbf{0.8847$\pm$0.021} & \textbf{0.8846$\pm$0.023} & \textbf{0.9450$\pm$0.010} & \textbf{0.7718$\pm$0.041} & \textbf{0.9390$\pm$0.010}\\
\hline
\end{tabular}}
\label{tab1}
\end{table}

% Additionally, to evaluate the statistical significance of the performance improvements, we conduct T-tests to assess performance improvement in \name. We adopt a significance level of $\alpha$ = 0.05. A p-value $\textless$ 0.05 provides strong evidence against the null hypothesis, indicating that the observed improvements in model performance are highly unlikely to result from random chance. As shown in Table~\ref{tab1}, our comprehensive statistical analysis reveals that \name outperforms most of methods with p-values $\textless$ 0.05, confirming that the performance gains are statistically significant. This rigorous significance testing offers robust evidence that the improvements achieved by \name over baseline methods are substantial, further validating its superior predictive capability in drug-target interaction tasks.

Additionally, to evaluate the statistical significance of performance improvements, we conduct t-tests for \name using a significance level of $\alpha$ = 0.05. A p-value less than 0.05 provides strong evidence against the null hypothesis, indicating that observed performance improvements are highly unlikely to result from random chance. As shown in Table~\ref{tab1}, statistical analysis confirms that \name outperforms most baseline methods with p-values below 0.05, demonstrating statistically significant gains. This rigorous testing provides robust evidence that improvements achieved by \name over baseline methods are substantial, further validating superior predictive capability in drug-target interaction tasks.

\subsection{Ablation study}
\name integrates three key designs to enhance DTI prediction performance: a domain-specific LLM encodes textual drug and target descriptions to introduce rich textual modalities; a dual cross-attention mechanism facilitates multi-modal alignment; the TSFusion module dynamically weights multi-modal data for effective fusion. To evaluate individual component contributions to \name, we conduct ablation experiments through systematic removal of the specific component: (1) \textit{\underline{w/o LLM\_Text}}: this variant replaces textual descriptions encoded by LLM with language models (e.g. Bert) encoding drug SMILES and protein amino acid sequences; (2) \textit{\underline{w/o Cra}}: this variant replaces dual cross-attention mechanism between modalities with self-attention confined to a single modality (3) \textit{\underline{w/o TSFusion}}: this variant replaces the TSFusion module with a static weight value (e.g. 0.5).

\begin{figure}[tp!]
\centering
{\includegraphics[width=\columnwidth]{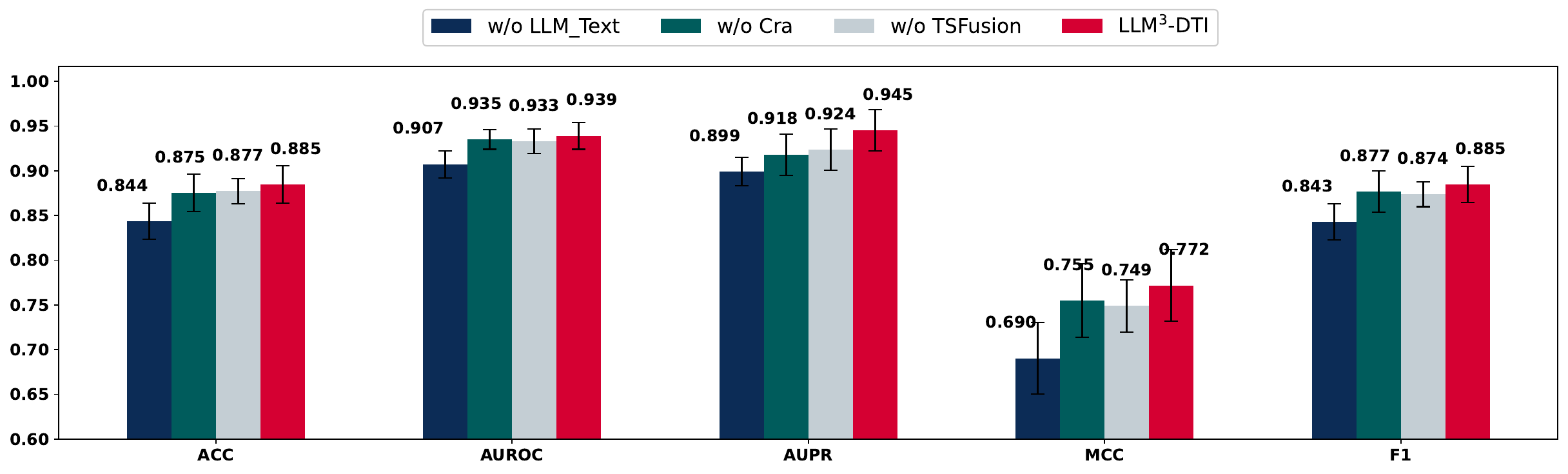}}
\caption{\gqh{Ablation study results.}}
\label{abla}
\end{figure}

Figure~\ref{abla} illustrates model performance following the ablation of specific modules. It can be observed that the removal of any module adversely affects DTI performance. First, Variant $\textit{w/o LLM\_Text}$ resulted in a significant decline across all metrics, indicating that text descriptions encoded by domain-specific LLM substantially enhance DTI task performance. Although language models encode SMILEs and amino acid sequences to introduce textual modalities, ablation experiments demonstrate that this approach inadequately exploits textual information. Second, Variants $\textit{w/o Cra}$ and $\textit{w/o TSFusion}$ exhibit modest declines across all metrics, indicating that cross-attention and dynamic weight allocation mechanisms between modalities are essential for further DTI performance improvement.

\subsection{Parameter sensitivity analysis.}\label{sec:pa}
% \lipsum[1]

\begin{figure}[bp!]
\centering
{\includegraphics[width=\columnwidth]{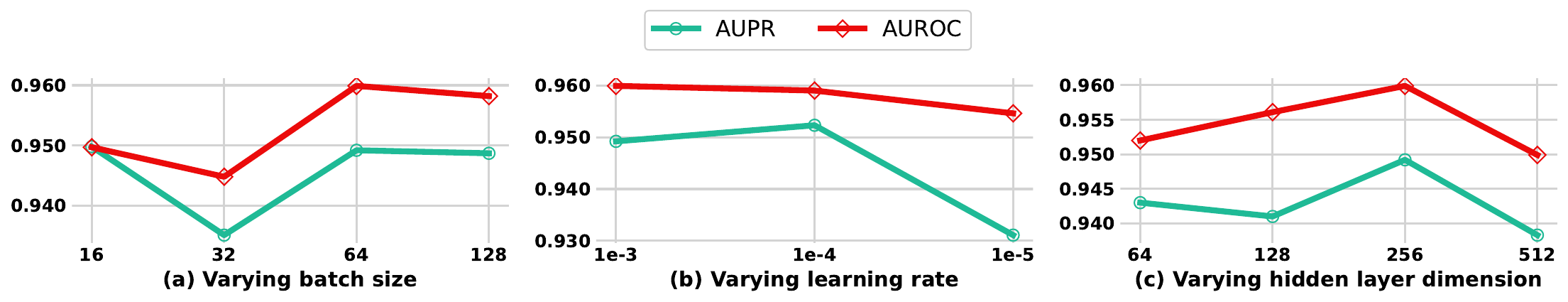}}
\caption{\gqh{Parameter sensitivity analysis.}}
\label{fig:pa}
\end{figure}

% \gqh{TBD}
In this section, we analyze the effect of some key parameters on \name performance: the batch size, the learning rate, and the hidden layer dimension. The results are shown in Figure~\ref{fig:pa}. We can find that \name maintains consistent excellent performance despite parameter variations. The specific conclusions are as follows. 

Firstly, the results of $\textit{\underline{varying batch size}}$ indicate that a small-size batch size reduces the sample richness, potentially causing insufficient sample discrimination and underfitting. Conversely, a larger size increases computational complexity and memory demands while hindering effective learning. Thus, a moderate batch size achieves an optimal balance between learning efficiency and model performance. 
Secondly, the results of $\textit{\underline{varying learning rate}}$ indicate that an excessively small learning rate may increase overfitting risk, degrade performance, and prolong training time by slowing convergence. Therefore, a learning rate of 1e-3 is implemented.
Finally, the results of $\textit{\underline{varying hidden layer dimension}}$ indicate that small hidden layer dimensions may limit feature representation, impairing model learnability. While the larger dimensions increase computational resource requirements and overfitting risk. Thus, we choose the dimensions 128 in our settings.

\section{Discussion}
% To further investigate the performance of \name, we conduct four targeted experiments: training with imbalanced data, analysis of cold-start scenarios, analysis of efficiency, and case studies. Detailed discussions of the results are presented in subsequent sections.

To further evaluate \name performance, we conduct four targeted experiments: imbalanced data training, cold-start scenario analysis, efficiency assessment, and case study. Additionally, we visualize embedding changes during \name training. Detailed discussions of the results are presented in subsequent sections.

\subsection{Imbalanced data training}
% In the context of DTI prediction, the training data provided to deep learning models is critical to their performance. The imbalance between the number of positive and negative samples in the dataset may impact the effectiveness of these models. To address this issue, experiments were conducted using imbalanced training data. During dataset construction, the negative sample size was set to be five times and ten times the size of the positive sample set. These setups were then compared with the performance of the second best model, IMCHGAN. Particular focus was given to the results of the AUPR and F1 metrics, as these are more appropriate for evaluating model performance under imbalanced data conditions. As illustrated in Figure~\ref{fig2}, although data imbalance may negatively affect model performance, the proposed model demonstrated better performance under both configurations compared to IMCHGAN. The \name model exhibited robust resistance to dataset imbalance, highlighting its strong adaptability under challenging conditions.

In the context of DTI prediction, training data quality significantly influences deep learning model performance. This section examines model behavior under imbalanced data conditions. Given the limited availability of positive samples in real-world scenarios, experiments employ training datasets with negative-to-positive sample ratios of 1:1, 5:1, and 10:1. The resulting model is benchmarked against the second-best baseline IMCHGAN. Since MCC and F1 scores effectively assess imbalanced data performance, these two metrics are prioritized for evaluation.

\begin{figure}[tp!]
\centering
\includegraphics[width=\textwidth]{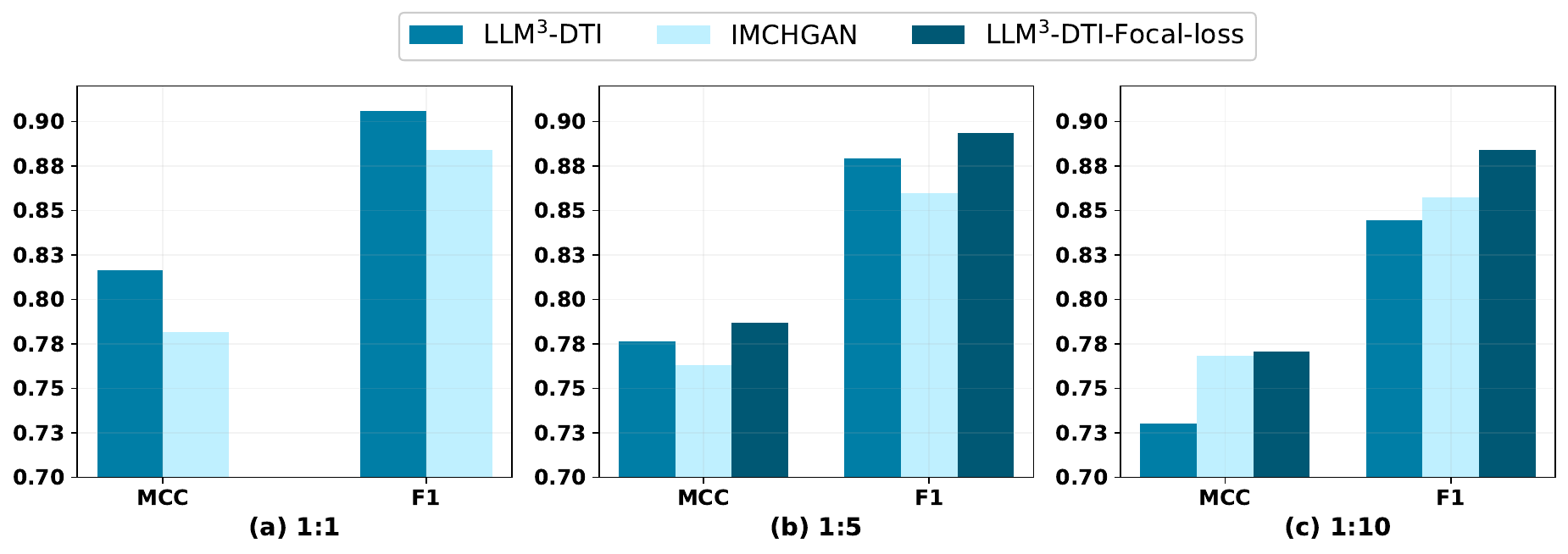}
\caption{\gqh{Imbalanced data training performance.}}
\label{imbalance}
\end{figure}

Analysis of Figure~\ref{imbalance} reveals two observations. First, increasing training set imbalance correlates with declining model performance, indicating that disproportionate positive-to-negative sample ratios adversely affect model efficacy. To ensure optimal performance, the model requires balanced training data. Second, while \name outperforms IMCHGAN at 1:1 and 1:5 positive-to-negative ratios, it exhibits greater performance degradation at a 1:10 ratio. Consequently, drawing on prior research,  we replace standard BCE loss with focal loss~\cite{lin2017focal} and retrain \name on imbalanced data, which enhances resistance to class imbalance and maintains robust performance under suboptimal conditions.

\begin{figure}[bp!]
\centering
\includegraphics[width=\columnwidth]{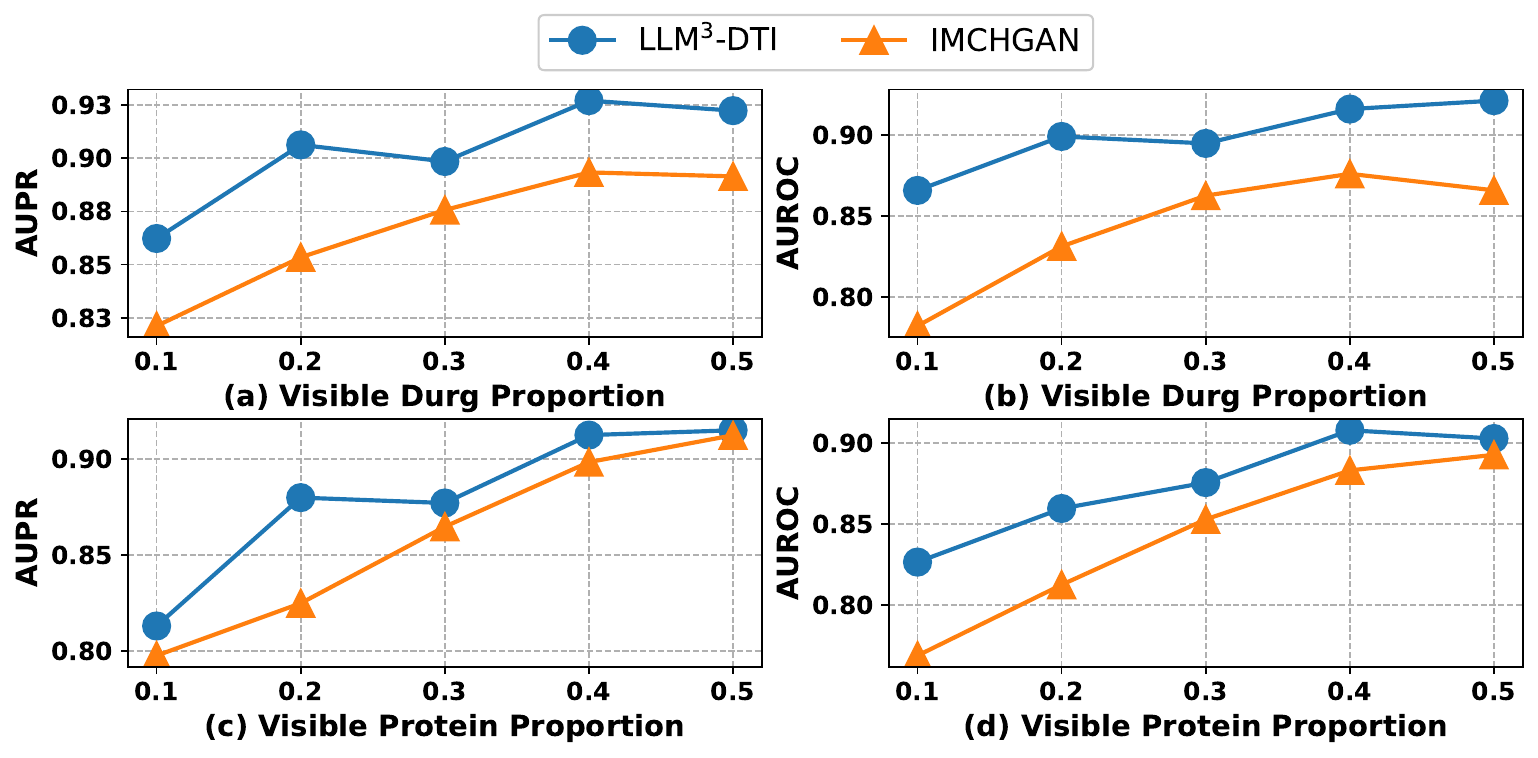}
\caption{\gqh{Cold start scenario performance.}}
\label{fig4}
\end{figure}

\subsection{Cold start scenario analysis}

To assess the generalization capacity of the proposed model, we conduct cold-start experiments, which partition training and test sets according to distinct drug and protein categories. We evaluate cold-start performance under two scenarios: drug cold-start and protein cold-start.
Specifically, we first set the proportion of drugs or proteins included in the training set. The model is subsequently evaluated on the DTI pairs comprising remaining drugs or proteins absent from the training set. For drug cold-start and protein cold-start scenarios, five experimental trials per scenario are conducted, with results compared against the second-best-performing baseline model.
This evaluates the ability to generalize to unseen drug or protein entities.

% As illustrated in Figures~\ref{fig3} and~\ref{fig4}, both \name and the second-best baseline, IMCHGAN, exhibit performance degradation under limited training data. However, \name consistently surpasses IMCHGAN across all configurations. Notably, \name demonstrates lower performance variance, whereas IMCHGAN displays pronounced instability as training data proportions vary. These results confirm \name’s superior robustness in cold-start scenarios.

Figure~\ref{fig4} illustrates the AUPR and AUROC results for drug and protein visibility proportion ranging from 0.1 to 0.5. We can find that model performance degrades when drug or protein visibility in the training set is limited. As the visibility proportion increases, the model performance improves progressively. Notably, \name outperforms IMCHGAN in all cold-start scenarios, indicating strong generalization capabilities and effectiveness in identifying DTIs absent during training.

% \begin{figure}[htp!]
% \centering
% \includegraphics[width=\textwidth]{figure/aupr_bigs.eps}
% \caption{Cold start experiments of drugs and proteins with different visible proportions of \name and IMCHGAN. The horizontal axis represents the different visible drugs or proteins proportion in the training set, while the vertical axis represents the values of the F1 metric.}
% \label{fig3}
% \end{figure}

% \begin{figure}[htp!]
% \centering
% {\includegraphics[width=\textwidth]{figure/f1_bigs.pdf}}
% \caption{Cold start experiments of drugs and proteins with different visible proportions of \name and IMCHGAN. The horizontal axis represents the different visible drugs or proteins proportion in the training set, while the vertical axis represents the values of the AUPR metric.}
% \label{fig4}
% \end{figure}

\subsection{Efficiency Assessment}
% When evaluating a model, convergence efficiency is an important consideration in addition to its performance. To mitigate the scale effects caused by different loss functions, the loss values were normalized to the range of 0 and 1 using min-max normalization. Figure~\ref{fig5} illustrates the loss curves of \name and IMCHGAN during the training process. An interesting phenomenon was observed. The number of training epochs required by \name to reach a stable loss value was significantly lower than that of IMCHGAN. This finding highlights a substantial advantage of lightweight design, which enables a relatively faster convergence while maintaining competitive performance.

% Additionally, apart from convergence speed, \name exhibited superior performance upon reaching a stable state compared to IMCHGAN. This suggests that the design and training methodology of \name facilitate a more effective capture of the underlying data characteristics, thereby enhancing model performance. These results underscore not only the benefits of the lightweight design in accelerating training but also the superiority of \name in handling specific tasks, providing strong evidence for its adoption and evaluation in practical applications. 

\begin{figure}[bp!]
  \centering
    \begin{minipage}[b]{\linewidth}
      \subfigure[Loss and performance evolution.]{
        \includegraphics[width=0.46\linewidth, height=0.3\linewidth]{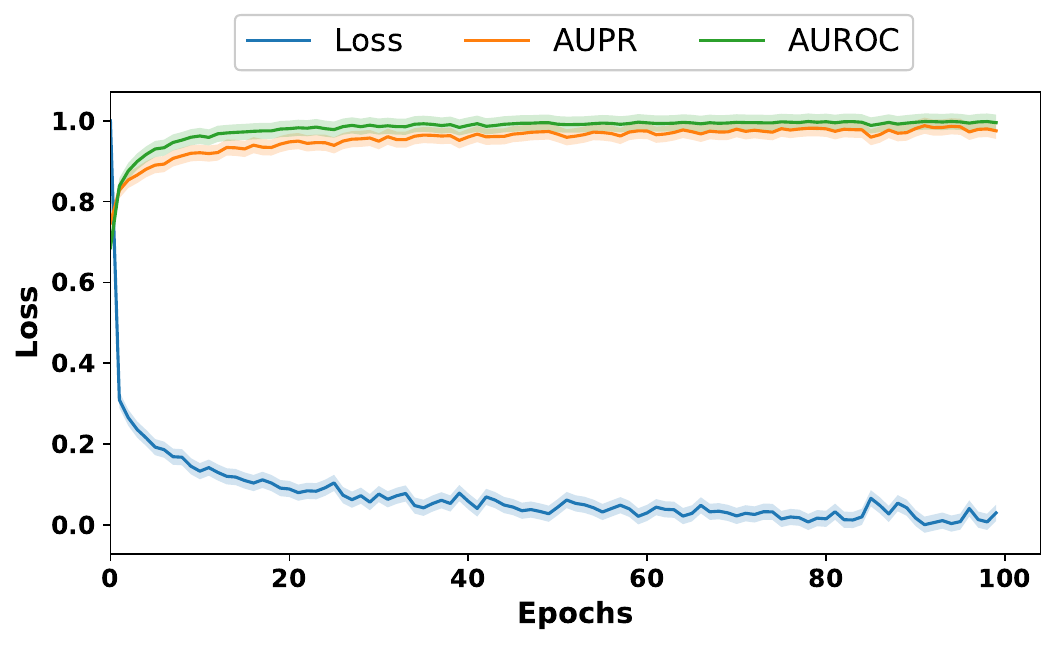}
        \label{ea-a}
      }
      \subfigure[Time and memory usage comparison.]{
        \includegraphics[width=0.46\linewidth, height=0.3\linewidth]{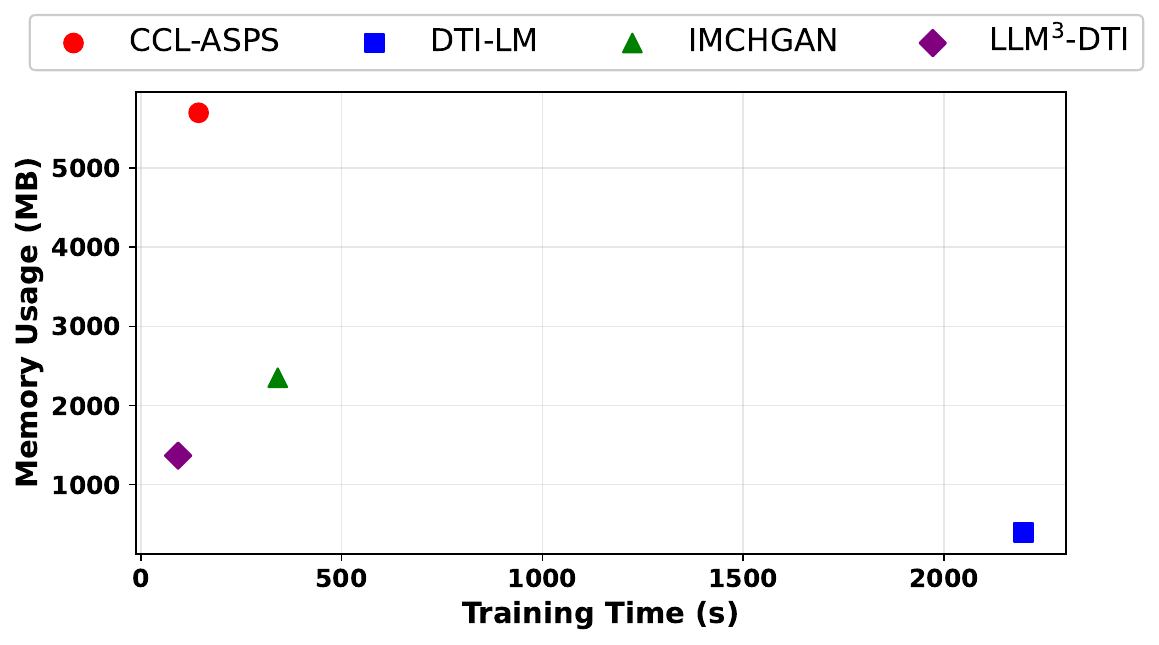}
        \label{ea-b}
      }
    \end{minipage}
    \caption{Efficiency assessment.}
    \label{fig:ea}
\end{figure}

The efficiency is also a critical evaluation criterion alongside model performance. Figure~\ref{fig:ea} reports the efficiency assessment of \name.
In Figure~\ref{ea-a}, \name achieves loss and performance stability in a few training epochs, demonstrating the efficiency of its architecture without performance degradation. Furthermore, \name attains superior performance at convergence. This implies that architecture and training strategy more effectively distill salient data patterns, enhancing predictive accuracy. These findings validate the dual advantages of design—accelerated convergence through abundant features and task-specific superiority—supporting its practicality in real-world applications.

Figure~\ref{ea-b} compares \name with some baselines in terms of training time and memory usage. It can be observed that CCL-ASPS requires more memory than other methods. Although DTI-LM consumes the least memory, it demands extended training time to achieve optimal performance. \name achieves the best predictive performance while requiring less training time and memory usage than most of the compared methods. This efficiency underscores the scalability of \name, making it well-suited for large-scale DTI prediction tasks.

% \begin{figure}[tp!]
% \centering
% {\includegraphics[width=0.5\textwidth, height=0.5\textwidth]{figure/loss_stack_big.pdf}}
% \caption{The normalized loss values of \name and IMCHGAN during the training process. \name can achieve convergence with fewer epochs.}
% \label{fig5}
% \end{figure}

% \begin{table}[tp!]
% \centering
% \caption{Comparison of model efficiency}
% \resizebox{\columnwidth}{!}{
% \begin{tabular}{lccc}
% \hline
% \textbf{Model} & \textbf{Training Time (s)} & \textbf{Memory Usage (MB)} & \textbf{Model Size (MB)} \\
% \hline
% CCL-ASPS        & 144                      & 5699                    & 72.54                 \\
% DTI-LM         & 2198                     & 392                     & 12.47                 \\
% IMCHGAN        & 341                      & 2351                    & 1.69                  \\
% \name         & 93                       & 1366                    & 47.07                 \\
% \hline
% \end{tabular}
% }
% \label{tab:model_comparison}
% \end{table}

\subsection{Case study}
Diabetes mellitus is a highly prevalent chronic metabolic disease globally, with incidence demonstrating a significant increase over recent decades, presenting a major public health challenge. To evaluate model reliability and practical applicability, we conduct a case study. Specifically, we first exclude those associated with diabetes-related drugs or targets from all DTI pairs. Then we train the \name on the remaining DTI data and evaluate it on the diabetes-specific DTI pair set.

Table~\ref{tab:cs} presents the top 10 confidence scores for model predictions in DTI with and without correlation analysis. We can find that \name demonstrates high accuracy in predicting novel drugs that are invisible in its training process. These results indicate that \name effectively screens and identifies promising candidates for further investigation.

% \gqh{TBD} To rigorously assess the reliability and practical applicability of the proposed model, a case study evaluates its performance. Drugs excluded from the training phase are selected to ensure predictions of potential DTIs rely on inferred intrinsic features rather than training data memorization or overfitting. The trained \name model scores binding affinities between these drugs and a protein set, generating interaction likelihoods for all pairs.

% Candidate predictions are ranked by model scores. The top 10 ranked drug-target pairs with high binding propensity were subjected to in-depth analysis. Biological relevance is validated through consultation of scientific literature, databases (e.g., DrugBank, BindingDB), and published experimental data. Validation criteria assess whether interactions are explicitly documented or reported as biologically significant. A parallel analysis focuses on protein targets.

% As summarized in Tables~\ref{tab3} and~\ref{tab4}, the 20 analyzed drug-target pairs under varied scenarios reveal that a significant proportion exhibit experimentally validated binding. These results demonstrate the capability of the model to reliably identify biologically associated drug-target pairs.

\begin{table}[tp!]
\centering
\caption{Case study of diabetes mellitus.}
\begin{tabular}{ccccc}
\hline
Drug Id & Protein Id & Ground Truth & Prediction & Correctness \\ \hline
DB00573          & P23219              & 1                     & 1                   & TRUE                 \\ 
DB00749          & P23219              & 1                     & 1                   & TRUE                 \\ 
DB00461          & P23219              & 1                     & 1                   & TRUE                 \\ 
DB01069          & P08173              & 1                     & 1                   & TRUE                 \\ 
DB01149          & Q01959              & 1                     & 1                   & TRUE                 \\ 
DB01173          & P23975              & 1                     & 1                   & TRUE                 \\ 
DB01165          & P11388              & 1                     & 1                   & TRUE                 \\ 
DB01221          & P14416              & 1                     & 1                   & TRUE                 \\ 
DB01221          & Q8TCU5              & 1                     & 1                   & TRUE                 \\ 
DB00413          & P08913              & 1                     & 1                   & TRUE                 \\ \hline
DB00981          & P13639              & 0                     & 0                   & TRUE                 \\ 
DB00485          & O14788              & 0                     & 0                   & TRUE                 \\ 
DB00485          & P49821              & 0                     & 0                   & TRUE                 \\ 
DB00869          & Q9P2R7              & 0                     & 0                   & TRUE                 \\ 
DB00373          & P04075              & 0                     & 0                   & TRUE                 \\ 
DB00661          & P13051              & 0                     & 0                   & TRUE                 \\ 
DB00373          & Q8NFA2              & 0                     & 0                   & TRUE                 \\ 
DB00621          & P04062              & 0                     & 0                   & TRUE                 \\
DB00549          & Q9Y4W6   & 0                     & 0                   & TRUE                 \\
DB00973          & P15144              & 0                     & 1                   & FALSE                                 \\ \hline
\end{tabular}
\label{tab:cs}
\end{table}

\subsection{Visualization}

\begin{figure}[tp!]
\centering
\includegraphics[width=\linewidth]{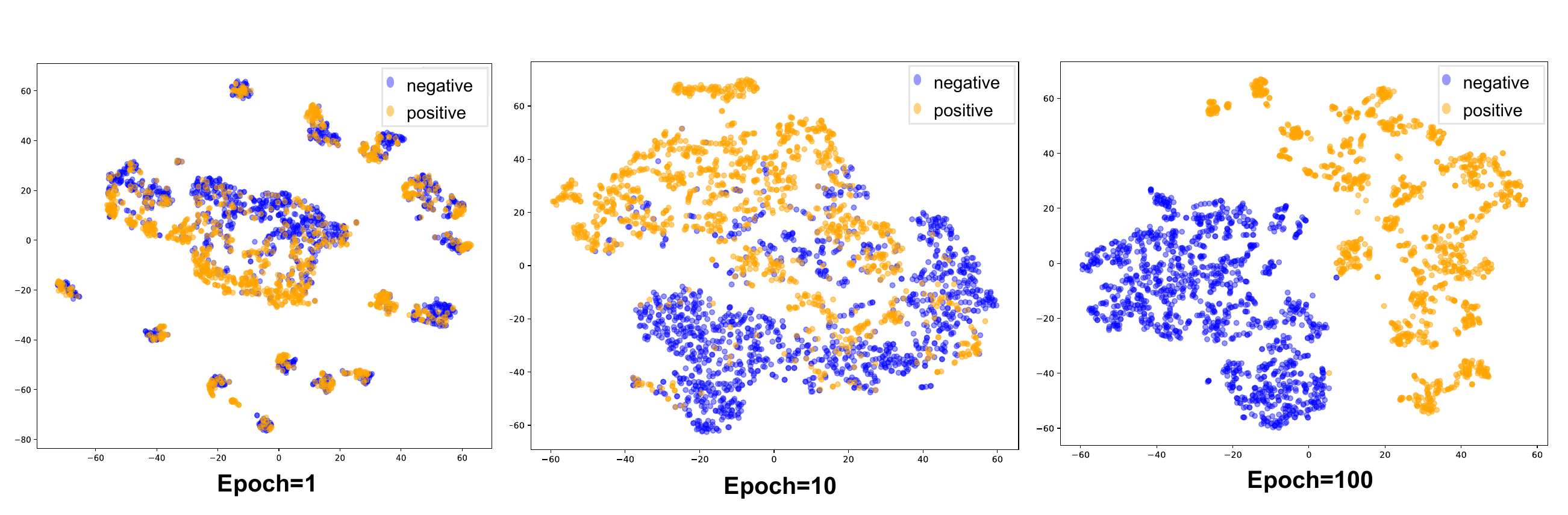}
\caption{Visualization results of positive and negative sample features during the training process. Orange and blue represent the mapping of positive and negative sample features in two-dimensional space, respectively. As the number of training epochs increases, positive and negative samples are significantly separated.}
\label{fig6}
\end{figure}

To intuitively demonstrate model performance and understand positional relationships within the latent space, we project the drug-target representations learned by \name into a two-dimensional space employing the t-distributed stochastic neighbor embedding algorithm (t-SNE). This dimensionality reduction technique visualizes clustering and separation of positive and negative samples during training. Figure~\ref{fig6} reveals that there is a substantial initial overlap between positive and negative sample representations, exhibiting unclear segregation. As training progresses, the representations gradually separate, forming distinct clusters aligned with true interaction labels. These visualizations confirm that \name successfully captures discriminative features and binding patterns between drugs and targets.

% \section{FUTURE WORK}
% The performance of the LLM in extracting meaningful features depends on the richness and accuracy of textual information from databases like DrugBank and UniProt. Incomplete or outdated data, especially for newly discovered or less-studied entities, may result in less informative embeddings and impact prediction accuracy. Additionally, biases or gaps in the LLM's training data can affect its ability to capture underrepresented drug or target mechanisms. The current model emphasizes general drug properties and target functions but does not incorporate other crucial factors, such as structural data, binding site details, or experimental conditions, which could further enhance predictions. Future work will address these limitations by integrating additional data modalities, refining LLM capabilities, and exploring multi-modal information integration.

\section{Conclusion}
% This study primarily demonstrates that leveraging rich textual information, as opposed to simple sequence data, along with the in depth understanding, modeling and iterative refinement of such features, significantly enhances the performance of DTI prediction tasks. Compared to using only SMILES strings for drugs and amino acid sequences for proteins, descriptive textual data for drugs and proteins encompass more comprehensive information, which better facilitates the determination of their binding relationships. While traditional language models struggle to interpret complex textual data, the advent of LLMs has successfully addressed this challenge. By introducing complex textual feature representations encoded by an LLM and updating them via a cross attention mechanism, \name effectively captures the salient features of drugs and proteins. Extensive experimental results demonstrate that the proposed approach achieved significant improvements in predictive accuracy compared with existing methods. \name not only serves as a powerful tool for drug repurposing but also presents a novel paradigm for future advancements in DTI prediction tasks. Moving forward, plans include integrating additional heterogeneous attributes of drugs and proteins and further validating the results through wet lab experiments.

This study demonstrates that leveraging rich textual information, rather than simplistic sequence data, can significantly enhance DTI prediction performance. Compared to relying solely on SMILES strings for drugs and amino acid sequences for proteins, descriptive textual data provides richer contextual information, which improves the inference of binding relationships. While traditional language models fail to interpret complex textual data, LLMs overcome this limitation. By integrating LLM-encoded textual features and refining multi-modal embeddings through the dual cross-attention and dynamically gate weighting fusion mechanisms, \name effectively captures discriminative drug and protein characteristics. Experimental results confirm that the proposed method achieves superior predictive accuracy compared to existing approaches. \name not only serves as a robust tool for drug repurposing but also establishes a novel framework for advancing DTI prediction methodologies. Future work will integrate additional heterogeneous attributes of drugs and proteins and validate predictions through wet-lab experimentation.

\section{Limitations and Future works}
The performance of the LLM in extracting meaningful features depends on the richness and accuracy of textual information from databases like DrugBank and UniProt. Incomplete or outdated data, especially for newly discovered or less-studied entities, may result in less informative embeddings and impact prediction accuracy. Biases or gaps in the training data of LLM can affect its ability to capture underrepresented drug or target mechanisms.  Additionally, more datasets containing a wider range of drugs and proteins should be considered for collection and use.
% The current model emphasizes general drug properties and target functions but does not incorporate other crucial factors, such as structural data, binding site details, or experimental conditions, which could further enhance predictions. 
Future work should consider integrating additional data modalities, refining LLM capabilities, and exploring multi-modal information integration.

\section{Data availability}
Codes and datasets are available at https://github.com/chaser-gua/LLM3DTI.

% \section{FUNDING}
% This work was supported by the National Key Research and Development Program [grant number 2022YFD1700200]; and the National Natural Science Foundation of China [grant number 32171330].

% \section{CRediT authorship contribution statement}
% \textbf{Qinghong Guo}: Methodology, Writing – original draft, Writing – review \& editing. \textbf{Yuhao Zhang}: Conceptualization, Data curation, Writing – original draft, Visualization.

% \section{Author contributions}
% Yuhao Zhang contributed to conceptualization, data curation, investigation, methodology, software development, visualization, and writing original draft. Qinghong Guo participated in conceptualization, data curation, investigation, methodology, software development, visualization, and writing original draft. Qixian Chen led conceptualization, methodology, project administration, resource acquisition, funding acquisition, and writing review and editing. Liuwei Zhang was responsible for formal analysis, investigation, resource acquisition, and validation. Hongyan Cui contributed to formal analysis, supervision, and writing review and editing, with a research focus on bioinformatics and novel therapeutic discovery. Xiyi Chen was involved in validation and writing review and editing.

\section{Acknowledgment}
The authors thank the reviewers for their valuable comments.

\section*{CRediT Author Contributions}
\textbf{Yuhao Zhang:} Conceptualization; Methodology; Investigation; Data Curation; Formal Analysis; Visualization; Writing – Original Draft.

\textbf{Qinghong Guo:} Conceptualization; Methodology; Investigation; Formal Analysis; Validation; Visualization; Writing – Original Draft.

\textbf{Qixian Chen:} Conceptualization; Resources; Supervision; Project Administration; Funding Acquisition; Writing – Review \& Editing.

\textbf{Liuwei Zhang:} Investigation; Data Curation; Software; Formal Analysis; Validation; Visualization.

\textbf{Hongyan Cui:} Investigation; Resources; Validation; Data Curation.

\textbf{Xiyi Chen:} Conceptualization; Resources; Supervision; Project Administration; Funding Acquisition; Writing – Review \& Editing.

\vspace{0.5em}

 \bibliographystyle{elsarticle-num} 
 \bibliography{myreference}





% \end{thebibliography}
\end{document}